\documentclass[3p,twocolumn]{elsarticle}
\usepackage[utf8]{inputenc}
\usepackage{amsmath}
\usepackage{amssymb}
\usepackage{amsthm}
\usepackage{graphicx}
\usepackage{multicol}
\usepackage[linesnumbered,ruled,vlined,commentsnumbered]{algorithm2e}
\SetAlFnt{\scriptsize}
\usepackage{dblfloatfix}
\usepackage{mathtools}
\usepackage{appendix}
\usepackage{balance}
\let\oldnl\nl
\newcommand{\nonl}{\renewcommand{\nl}{\let\nl\oldnl}}

\journal{Robotics and Autonomous Systems}

\begin{document}

\begin{frontmatter}

\title{Dynamic Prioritization for Conflict-Free Path Planning of Multi-Robot Systems}
\author[1]{Aditya Rathi}
\ead{rathi.aditya@alumni.iitgn.ac.in}

\author[2]{Rohith G}
\ead{rohith044@gmail.com}

\author[3]{Madhu Vadali\corref{correspondingauthor}}
\ead{madhu.vadali@iitgn.ac.in}
\cortext[correspondingauthor]{Corresponding author}
\date{January 2021}

\address{SMART Lab, Indian Institute of Technology Gandhinagar, Gandhinagar, Gujarat 382355, India}

\begin{abstract}
Planning collision-free paths for multi-robot systems (MRS) is a challenging problem because of the safety and efficiency constraints required for real-world solutions. Even though coupled path planning approaches provide optimal collision-free paths for each agent of the MRS, they search the composite space of all the agents and therefore, suffer from exponential increase in computation with the number of robots. On the other hand,  prioritized approaches provide a practical solution to applications with large number of robots, especially when path computation time and collision avoidance take precedence over guaranteed globally optimal solution. While most centrally-planned algorithms use static prioritization, a dynamic prioritization algorithm, PD*, is proposed that employs a novel metric, called freedom index, to decide the priority order of the robots at each time step. This allows the PD* algorithm to simultaneously plan the next step for all robots while ensuring collision-free operation in obstacle ridden environments. Extensive simulations were performed to test and compare the performance of the proposed PD* scheme with other state-of-the-art algorithms. It was found that PD* improves upon the computational time by 25\% while providing solutions of similar path lengths. Increase in efficiency was particularly prominent in scenarios with large number of robots and/or higher obstacle densities, where the probability of collisions is higher, suggesting the suitability of PD* in solving such problems.   

\end{abstract}
\begin{keyword}
Collision-free paths, Dynamic prioritization, Multi-robot systems, Path planning, D* Lite.  
\end{keyword}

\end{frontmatter}

\section{Introduction}\label{Introduction}
Advancements in technology has led to the widespread acceptance of robotic systems, especially in industrial environments \cite{acemoglu2020robots}. Moreover, in recent years, large scale deployment of robots, in the form of multi-robot systems, for various tasks such as the delivery of goods \cite{inproceedings}, warehouse management \cite{enright_optimization,arbanas2018decentralized}, cooperative driving \cite{hyldmar2019fleet,tsugawa_transport},  or mobility-on-demand systems \cite{Pavone_load_balancing} are becoming increasingly popular and viable. Multi-robot systems (MRS) portray emergent properties as they can effectively share both workload and functionality \cite{parker2007distributed}. This allows MRS to effectively deal with challenging tasks without making complicated modifications/interventions to an individual robot design.

However, MRS continue to suffer from several problems and challenges. Except for systems with a central planner, MRS make collective decisions. For example, \cite{olcay2020collective} describes a collective navigation strategy while \cite{de2019bio} explains a bio-inspired cooperative exploration task. Both works highlight the inherent coordination required for a successful MRS wherein each robot must account for multiple other robots before making a decision. Depending on the situation, further complexity may be added in the form of dynamic team sizes and interactions, heterogeneity in terms of either hardware or software differences, and specific constraints imposed by the task at hand \cite{Geihs_2020}. This makes the problems that are trivial for single robots significantly more complex for MRS.

\begin{figure}[h!]
    \centering
    \includegraphics[width=\linewidth]{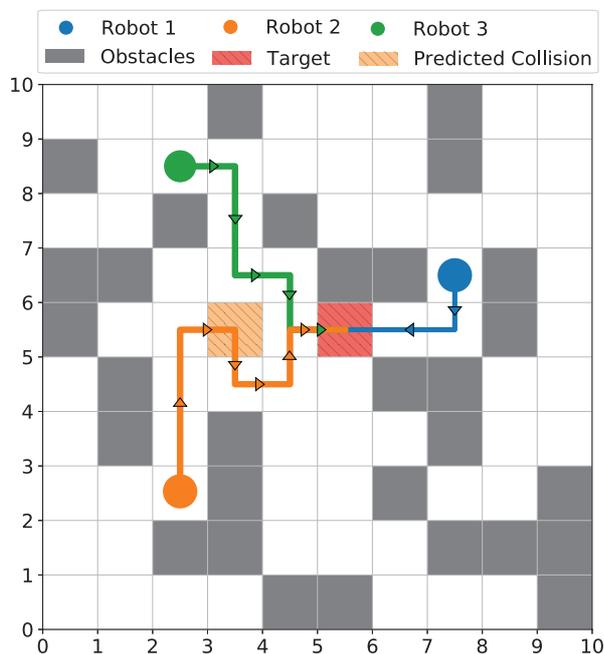}
    \caption{An illustrative example of MRS path planning problem}
    \label{fig:motivation_field}
\end{figure}

Path planning, which is a fundamental task in robotics, faces similar issues when expanding to MRS. Typically computing a path for one robot from one point to another in a given environment is not difficult. Multiple algorithms, either heuristics-based (like A* \cite{nilsson2014principles}, D* Lite \cite{koenig2002d}, Djikstra’s algorithm \cite{dijkstra1959note}) or otherwise (such as breadth-first search \cite{moore1959shortest} or depth-first search \cite{tarjan1972depth}), exist for this task of planning the path for a single robot. On the other hand, in MRS, multiple robots share the environment (typically unstructured) in which they operate. Thus, while planning paths for these systems, it is imperative not only to ensure that the robots can efficiently reach their targets but also to do so in a safe and collision-free manner. Therefore, cooperation and coordination among individual agents becomes an essential criterion for solving this problem. An illustrative example of this problem is shown in Figure \ref{fig:motivation_field} wherein the environment is modelled as a 10$\times$10 grid map. The MRS consists of three robots, at different initial positions, trying to reach one target. It can be seen that the two robots, $R_2$ starting at $(2,2)$ and $R_3$ starting at $(2,8)$, will have a collision at $(3,5)$. Thus, the algorithm that will solve this problem must be capable of preemptively handling this collision. A possible solution is shown in Figure \ref{fig:motivation_field}, where $R_2$ has to take a detour to avoid a collision with $R_3$.

Furthermore, rather than merely planning a safe and collision-free path for each agent, it is essential to find an optimal path, preferably a global optimum. A path is said to be optimal if the sum of its transition costs (edge costs) is minimal across all possible paths from the initial position (start state) to the goal/final position (goal state) \cite{ferguson2005guide}. For MRS, the solution is globally optimal if the sum of edge costs is minimal over every robot’s possible paths. As the number of robots in the system increases, the possible path options that need to be considered also increase. Naturally, this leads to an exponential increase in computational complexity with the increase in number of robots \cite{lavalle2006planning}. Meeting these targets of safety and optimality makes the MRS path planning task significantly more challenging than computing the path for a single robot. 

Approaches for solving the multi-robot path planning problem can be broadly classified into two categories:
\begin{itemize}
    \item \textit{Coordinated or coupled approach} in which the path is computed in the composite configuration space of all the robots
    \item \textit{Prioritized or decoupled approach} in which the path for robots are computed sequentially following a particular priority order.
\end{itemize}

A significant amount of research has been conducted on the coordinated path planning approach for MRS \cite{barraquand1991robot,olmi2011efficient,draganjac2016decentralized}. These approaches solve for the path in the composite configuration space of all the robots. This composite space is obtained through the Cartesian product of the configurations of all the robots. The solutions generated typically guarantee a global optimum. However, as these methods operate in the composite configuration space, they require large computational resources. Despite efforts in reducing the computational load of these methods \cite{ferner2013odrm,wagner2011m}, they still scale poorly to environments with a large number of robots.

On the other hand, prioritized planning has been studied less extensively even though they offer much more practically viable solutions to the MRS path planning problem. In these approaches, the robots’ paths are planned sequentially using a priority scheme, effectively deconstructing the MRS path planning problem into a series of single robot path planning problems. Robots with a higher priority get their paths planned first and have more options in their path choices. Robots with a lower priority then adapt their paths to ensure that they do not collide with higher priority robots. These methods can be further classified into \textit{centralized} and \textit{decentralized} approaches. In the centralized approach, a central planner plans the priorities and paths for all the robots, while in the decentralized approach, the robots plan their paths collectively through cooperation and coordination.

The quality of the solutions obtained through prioritized planning depends on the choice of the prioritization scheme, particularly in restricted environments with limited path choices \cite{van2005prioritized}. Prioritized solutions that have been proposed generally use a static priority scheme (for example \cite{jose2016task}), i.e., the priority to all the robots are assigned at the start based on some criterion and then this priority is used throughout the problem. This can potentially lead to the priority order being sub-optimal at some time steps implying that a different priority order exists at these time steps which would lead to a more efficient path. Further, these approaches plan the path of one robot completely before planning the next robot’s path \cite{van2005prioritized,bennewitz2002finding}. This is not desired because modern systems have sufficient computational resources to handle multiple robots simultaneously. A simultaneous approach can possibly provide benefits in terms of computational efficiency.

In this paper, a new dynamic prioritization method, called PD*, for MRS path planning is proposed to compute the priorities of each robot at every time step. To assign the priority of a robot, a new metric, called the freedom index, is defined. As the priorities of the robots are dynamically changed, the optimal priority order is ensured at every time step. The dynamic priorities are further utilized to simultaneously compute the immediate next-step path of all the robots. This method incorporates the solution quality improvements of a coupled planning approach with the computational efficiency improvements of a prioritized approach. Extensive simulations have been performed to test the effectiveness of the proposed PD* algorithm and compare its performance with existing state-of-the-art algorithms. Results have been reported in terms of both the solution quality (in terms of path length) and the computational time taken by the solutions. It will be shown that the algorithm provides solutions of approximately equal quality with significant improvements in terms of computational efficiency.

The rest of this paper is organized as follows: Section II states the problem and the proposed PD* algorithm. Section III discusses simulation results and comparisons with other algorithms from literature. Finally, Section IV concludes the paper. 

\section{Theory}\label{Theory}
The task addressed in this paper is the path planning of a MRS in an unstructured environment. In particular, all the robots have different starting points but have to move to a common goal or target. The following sections formalize the problem and explain the proposed PD* algorithm.

\subsection{Problem Description}\label{Prob_desc}
Consider a multi-robot system with $n$ robots $R_1, R_2,\ldots, R_n$ in a two-dimensional environment. All the robots are identical in their geometry and performance. For each robot, a starting configuration $r_i$ and the common goal (target) $G$ are defined. The task is to find a collision-free path $P_i$ for each robot $R_i$ through the given environment such that $P_i(0) = r_i$ and $P_i(T_i) = G$ where $T_i$ is the time when robot $R_i$ reaches the goal.

A ‘no stopping’ condition is also imposed on the robots preventing them from waiting at a given location to avoid collisions. Further, it is assumed that $r_i \neq r_j \forall \text{ } i,j$, and each robot moves at a constant speed over its path $P_i$. The path thus calculated also needs to account for the static obstacles present in the environment as well as ensure that there is no collision between any two robots i.e. $P_i(t) \neq P_j(t) \forall \text{ } i,j,t$.

\subsubsection{Discretization of Environment}\label{descretize}
\begin{figure}[h!]
    \centering
    \includegraphics[width=\linewidth]{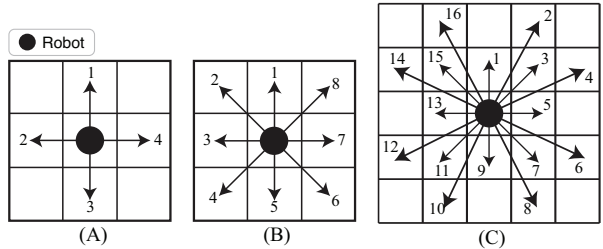}
    \caption{Types of edge connections for (A) 4-connected grid world, (B) 8-connected grid world, and (C) 16-connected grid world}
    \label{fig:connectedness}
\end{figure}

The robots are in a continuous time and space configuration in the real world, which is discretized for solving the path planning problem. While multiple approaches have been used to discretize the terrain \cite{zimmer2020adaptive,lozano,henrichcspace} a popular method is to split the terrain into squares of equal size\cite{erssonunknown,stentz1997optimal}. This creates a grid world, where each square is a node with transition edges connecting them. Depending on the environment the configuration of the transition edges may vary. While 4 and 8-connected grid worlds are common, more complex representations are also possible as shown in Figure \ref{fig:connectedness}. Each transition is assigned a cost according to the cost function defined by the problem. This combination of the grid world nodes and the transition edges form a graph. 

Without loss of generality, a 4-connected discrete occupancy graph is used in this paper. Discrete occupancy implies that at most one object (robot/obstacle) can occupy a node. Further, the transition cost function used is the distance between the nodes. Thus, for a 4-connected graph the transition cost of all neighbors of a node are equal. Mathematically, let $S$ denote the finite set of nodes in the graph and individual nodes be represented by $s$. Thus $P_i(j) \in S~\forall ~i,j$.

 Time is discretely represented in the form of time steps such that the time interval, $\Delta t$, between two time steps is equal to the time taken by a robot to move from one node to another. Thus, $T_i$ now represents the time step in which $R_i$ reaches $G$. Finally, the transition cost of moving from node $s$ to $s'$ for some robot $R_i$ is denoted by $c(s,s')$. For neighboring nodes $s$ and $s'$, $c(s,s') \vcentcolon = 1$ 

\subsection{Path Planning Algorithm}\label{path_algo}
The goal of a path planning algorithm is to find an efficient path to move a robot from the starting node to the target node while avoiding all obstacles in the environment. Since the prioritized approach decomposes the MRS path planning problem into a series of single robot path planning problems we need a suitable algorithm for this latter task. In this regard, heuristic based approaches are preferred because they can prune the search space, significantly speeding up the process. These approaches have an inherent metric to differentiate between the various possible elements of $S$ to find the ideal set of vertices to reach the goal. Thus they can focus their search on potentially favorable vertices leading to a faster solution. 

A wide variety of path planning algorithms are available for the path planning task, viz., A* \cite{nilsson2014principles}, D* \cite{D*}, D* Lite \cite{koenig2002d}, ARA* \cite{ARA*}. Of these, the A* algorithm is popular in MRS path planning problems. However, in proposed method, D* Lite is used as the base path planning algorithm as it provides several advantages over A*. These will be discussed in the following section. 

As the base path planning algorithm, D* Lite handles the task of finding the shortest path for every robot in the given environment. Naturally, D* Lite is designed for single robot path planning and needs to be extended for MRS. Furthermore, the proposed algorithm is not limited to D* Lite and can be replaced with any other path planning technique.

\subsubsection{Advantages of using D* Lite over A*}\label{path_algo_adv}
It is important to address the rationalization behind the choice of using D* Lite over the A* algorithm. For the class of problems addressed in this paper, D* Lite provides two distinct advantages over A*.

A* searches for the shortest path from the robot’s starting position to the goal, whereas D* Lite searches in a reverse direction, i.e., from the goal to the starting position. In the problem being addressed in this paper, it would be necessary to re-plan the path of some robots to avoid collisions. At this point, the start position of the robots would have changed, but the target position remains fixed throughout the motion. Thus, distances of nodes from the target would likely remain unchanged whereas distance calculated from the starting position would be different. 

This naturally leads to the second advantage that D* Lite provides over A*. When there is a change in the environment (an obstacle is added or removed), D* Lite can reuse old information by smartly updating only the vertices of the graph affected by the change. On the other hand A* has to recalculate the entire path. Thus, D* Lite can handle re-plans significantly faster than A*.

In the problem described in Section \ref{Prob_desc}, the goal location is fixed while the robots are moving and virtual obstacles are added to mitigate collisions. This makes the environment dynamic resulting in the robots having to recalculate paths multiple times to mitigate collision. In this scenario D* Lite provides superior performance compared to A* as it has to do minimal recalculation to update the paths every time the graph changes. This increase in computational efficiency of D* Lite over A* has also been replicated in other experiments \cite{setiawan2014experimental}, \cite{Rosa_Comp}.

\subsubsection{D* Lite}\label{D*_lite}
\begin{algorithm*}[!hb]
\begin{multicols}{2}
\DontPrintSemicolon
\SetKwFor{While}{while}{:}{}
\SetKwFor{For}{for}{:}{}
\SetKwIF{If}{ElseIf}{Else}{if}{:}{else if}{else:}{}
\SetKwFunction{CalcKey}{CalculateKey}
\SetKwProg{CK}{Function}{:}{}
\CK{\CalcKey{s}}{
\KwRet $[min(g(s),rhs(s)) + h(s_{start},s)+k_m; min(g(s),rhs(s))]$\;
}

\SetKwFunction{init}{Initialize}
\SetKwProg{IN}{Function}{:}{}
\IN{\init{}}{
$U = \emptyset$\;
$k_m = 0$\;
for all $s\in S: rhs(s)=g(s)=\infty$\;
$rhs(s_{goal}) = 0$\;
$U.insert(s_{goal}$,\CalcKey{$s_{goal}$})\;
}

\SetKwFunction{updatevertex}{UpdateVertex}
\SetKwProg{UP}{Function}{:}{}
\UP{\updatevertex{u}}{
if $(u \neq s_{goal})$: $rhs(u) = min_{s'\in Succ(u)}(c(u,s')+g(s'))$\;
if $u \in U$: U.Remove(u)\;
if $(g(u) \neq rhs(u))$: U.Insert(u, \CalcKey{u})\;
}

\SetKwFunction{compsp}{ComputeShortestPath}
\SetKwProg{CSP}{Function}{:}{}
\CSP{\compsp{}}{
\While{U.TopKey() $<$ \CalcKey{$s_{start}$} OR $rhs_{s_{start}} \neq g(s_{start})$}{
$k_{old}$ = U.TopKey()\;
u=U.Pop()\;
\If{$k_{old}<$\CalcKey{u}}{
U.Insert(u,\CalcKey{u})\;}
\ElseIf{$g(u)>rhs(u)$}{
$g(u) = rhs(u)$\;
for all $s \in Pred(u)$: \updatevertex{s}\;}
\Else{
$g(u) = \infty$\;
for all $s \in Pred(u)\cup \{u\}$: \updatevertex{s}\;
}}}

\SetKwFunction{main}{Main}
\SetKwProg{MN}{Function}{:}{}
\MN{\main{}}{
$s_{last} = s_{start}$\;
\init{}\;
\compsp{}\;
\While {$s_{start} \neq s_{goal}$}{ 
if ($g(s_{start}=\infty$): then there is no known path\;
$s_{start} =$arg $min_{s'\in Succ(s_{start})}(c(s_{start},s')+g(s'))$\;
Move to $s_{start}$\;
Scan graph for changed edge costs\;
\If {any edge costs changed}{
$k_m = k_m + h(s_{last},s_{start})$\;
$s_{last}=s_{start}$\;
\For{all directed edges $(u,v)$ with changed edge costs}{
Update the edge cost $c(u,v)$\;
\updatevertex{u}\;}
\compsp{}\;
}}}
\caption{D* Lite \cite{koenig2002d}}
\label{ogD*}
\end{multicols}
\vspace{2ex}
\end{algorithm*}

D* Lite uses a distance-based heuristic to find the shortest path from the start to the goal. The algorithm assigns $g$ values to every node $s$ which represent the cost incurred to reach the node $s$ from the target $G$. 

Initially the $g$ values of the nodes are unknown as the graph has not been explored. To assign these $g$ values the algorithm maintains a priority list. Nodes whose $g$ values have been updated are added to the priority list as possible points of further exploration. The list is sorted using the distance from the starting location. Nodes closest to the starting location are explored first and then removed from the priority list once that node has been fully explored.

The process starts from the target node and is repeated until the algorithm reaches the starting node. At this point the path is successfully planned. The algorithm then uses the $g$ values computed to find the shortest path `back' to the target. This heuristics based approach allows D* Lite to identify the shortest path without exploring all the nodes, improving efficiency.

When the graph changes, D* Lite uses the $g$ values to identify the nodes that are both affected by this change and are a part of the shortest path and only updates these nodes. To achieve this, it also leverages an `rhs' value which is the one-step look-ahead cost, and satisfies the following equation: 
\begin{equation}
    rhs(s)=
    \begin{cases}
    0 & \text{if } s=s_{goal}\\
    min_{s'\in Succ(s)}(f(s,s')) & \text{otherwise}
    \end{cases}
    \label{rhs}
\end{equation}

\noindent where $Succ(s)$ is the set of successor nodes of $s$, $f(s,s')=c(s,s')+g(s')$, and $c(s,s’)$ is the cost of reaching $s’$ from $s$ (the edge cost). By using both the $g(s')$ values and the $rhs(s)$ values, D* Lite is able to identify if the change in the graph has affected the node. This allows it to minimize the number of updates required to compute the new path, further improving efficiency.  

When the algorithm is first called, it initializes the graph and sets the $g$ values of all the nodes to infinity. The $rhs$ values are also calculated using the Eq.~(\ref{rhs}). The $rhs$ value of the goal is set to zero, and this node is added to the priority list. The algorithm then expands on the nodes until it reaches the start position. At this point the initial path is calculated. Now, the algorithm waits for a change in the graph and then updates the relevant nodes accordingly. Inconsistencies are settled and the updated path is found. 

Algorithm \ref{ogD*} shows the original D* Lite algorithm, as presented in \cite{koenig2002d}. 


\subsection{Prioritization}

The D* Lite algorithm must be extended to plan collision-free paths for each agent in the MRS. This requires a prioritization scheme. In this paper, the prioritization is based on a new metric called \textit{freedom index}. The following section defines the freedom index and explains its role in extending the D* Lite algorithm.

\subsubsection{The Freedom Index}

For a robot $R_i$ at a vertex $u$ the freedom index is defined as the number of possible vertices that $R_i$ can occupy in the next time step. Algorithm \ref{freedom_index_algo} shows the methodology to compute the Freedom index of a robot. Here, $Succ(u)$ is the set of all vertices that are the successors of $u$. It’s elements depend on the definition of the graph (as defined in section \ref{descretize}). If the graph is undirected, i.e. all the neighbors of a cell are indistinguishable from each other, then $Succ(u)$ will contain the neighbors of $u$ whereas if the graph is directed, meaning that the graph has a notion of `forward' and `backward,' then $Succ(u)$ will contain all the neighbors of u except the vertex that was previously occupied by $R_i$. $c[i](u,s)$ is the cost incurred by $R_i$ in moving from vertex $u$ to $s$. If this cost is infinity, then it implies that $s$ is a wall and thus this vertex will be excluded from the calculation of the freedom index.

The Freedom index of robot $R_i$ at vertex $u$ is a quantitative measure of the number of choices or the ‘freedom’ that $R_i$ has in choosing its next position. Robots with a higher freedom index have a larger number of alternatives to choose from in the case that they are unable to choose their first preference. Keeping in line with this observation, a robot $R_i$ with a smaller freedom index is given a higher priority than a robot $R_j$ with a larger freedom index. Ties, i.e., two robots $R_i$ and $R_j$ having equal freedom index, are arbitrarily broken when assigning priorities. 
\begin{algorithm}[h!]
\DontPrintSemicolon
\SetKwFor{While}{while}{:}{}
\SetKwFor{For}{for}{:}{}
\SetKwIF{If}{ElseIf}{Else}{if}{:}{else if}{else:}{}
\SetKwFunction{compf}{ComputeFreedom}
\SetKwProg{CF}{Function}{:}{}
\CF{\compf{i,u}}{
freedom\_index = 0\;
\For{$s\in Succ(u)$}{\If{$c[i](u,s)\neq \infty$}{$freedom\_index$+=1 \;}}
\KwRet freedom\_index
}
\caption{Freedom Index}
\label{freedom_index_algo}
\end{algorithm}
\vspace{-20 pt}
\subsubsection{Dynamic Prioritization}
The freedom index and the computed priority order is valid only for the current time step as it is dependent on the immediate neighborhoods of the individual robots. However, the definition of the freedom index creates a framework for efficiently calculating the priorities of all the robots at every time step.

This allows the proposed algorithm to calculate the priorities dynamically. Unlike a static priority ordering where the priorities may not be efficient at some time steps, the proposed prioritization results in an optimal hierarchy in the robots at every time step. Thus, the proposed algorithm is able to exploit the benefits of dynamic prioritization which is typically found in coupled approaches while maintaining the computational efficiency of a prioritized approach.

Furthermore, since the algorithm has knowledge of the priorities of the robots at every time step, it can plan this step for all the robots simultaneously. This is in stark contrast to traditional prioritized algorithms where the complete path of the robot with higher priority is planned before planning the path of lower priority robots.

\subsubsection{Collision Handling}

\begin{algorithm}[b!]
\DontPrintSemicolon
\SetKwFor{While}{while}{:}{}
\SetKwFor{For}{for}{:}{}
\SetKwIF{If}{ElseIf}{Else}{if}{:}{else if}{else:}{}
\SetKwFunction{resconf}{ResolveConflicts}
\SetKwProg{RCF}{Function}{:}{}
\SetKwFunction{try}{try}
\SetKwProg{tryy}{}{:}{}
\SetKwFunction{except}{except}
\SetKwProg{exc}{}{:}{}
\RCF{\resconf{SuccSteps,PriorityOrder}}{
PriorityOrder.Sort(key=FreedomIndex)\;
\For{idx $\in$ PriorityOrder}{
Check for collision in SuccSteps with all robots in PriorityOrder[0:idx]\;
\If{Collision Detected}{
\tryy{\try}{
g[idx'] = g[idx]\; 
rhs[idx'] = rhs[idx]\; 
U[idx'] = U[idx]\;
c[idx'](u,v) = c[idx](u,v) $\forall (u,v) \in S$\;
c[idx']($s[idx]_{start}$,SuccSteps[idx]) = $\infty$\;
\updatevertex{idx',SuccSteps[idx]}\;
\compsp{idx'}\;
SuccSteps[idx] = arg$min_{s'\in Succ(s[idx]_{start})} h(Z)$\;
}
\exc{\except {No Path}}{
SuccSteps[idx] = Previous Position of $R_{idx}$\;
}}}
\KwRet SuccSteps}
\nonl where, $Z = [s,s',idx,idx']$\\ 
\nonl and, $h(Z)= (c[idx'](s[idx]_{start},s')+g[idx'](s'))$
\caption{Conflict Resolution}
\label{conflict_res}
\end{algorithm}

\begin{algorithm*}[!b]
\begin{multicols}{2}
\DontPrintSemicolon
\SetKwFor{While}{while}{:}{}
\SetKwFor{For}{for}{:}{}
\SetKwIF{If}{ElseIf}{Else}{if}{:}{else if}{else:}{}

\SetKwFunction{CalcKey}{CalculateKey}
\SetKwProg{CK}{Function}{:}{}
\CK{\CalcKey{i,s}}{
\KwRet $[min(g[i](s),rhs[i](s)) + h(s_{start}[i],s)+k_m[i]; min(g[i](s),rhs[i](s))]$\;
}

\SetKwFunction{init}{Initialize}
\SetKwProg{IN}{Function}{:}{}
\IN{\init{i}}{
$U[i] = \emptyset$\;
$k_m[i] = 0$\;
for all $s\in S: rhs[i](s)=g[i](s)=\infty$\;
$rhs[i](s[i]_{goal}) = 0$\;
$U[i].insert(s[i]_{goal}$,\CalcKey{i,$s_{goal}$})\;
}

\SetKwFunction{updatevertex}{UpdateVertex}
\SetKwProg{UP}{Function}{:}{}
\UP{\updatevertex{i,u}}{
if $(u \neq s[i]_{goal}): rhs[i](u) = min_{s'\in Succ(u)}(c[i](u,s')+g[i](s'))$\;
if $u \in U[i]$: U[i].Remove(u)\;
if $(g[i](u) \neq rhs[i](u))$: U[i].Insert(u, \CalcKey{i,u})\;
}

\SetKwFunction{compsp}{ComputeShortestPath}
\SetKwProg{CSP}{Function}{:}{}
\CSP{\compsp{i}}{
\While{U[i].TopKey() $<$ \CalcKey{i,$s_{start}$} OR $rhs[i]_{s_{start}} \neq g[i](s_{start})$}{
$k_{old}$ = U[i].TopKey()\;
u=U[i].Pop()\;
\If{$k_{old}<$\CalcKey{i,u})}{
U[i].Insert(u,\CalcKey{i,u})\;}
\ElseIf{$g[i]_{u} > rhs[i](u)$}{
$g[i](u) = rhs[i](u)$\;
for all $s \in Pred(u)$: \updatevertex{i,s}\;}
\Else{
$g[i](u) = \infty$\;
for all $s \in Pred(u)\cup \{u\}$: \updatevertex{i,s}\;}}
}

\SetKwFunction{resconf}{ResolveConflicts}
\SetKwProg{RCF}{Function}{:}{}
\SetKwFunction{try}{try}
\SetKwProg{tryy}{}{:}{}
\SetKwFunction{except}{except}
\SetKwProg{exc}{}{:}{}
\RCF{\resconf{SuccSteps,PriorityOrder}}{
PriorityOrder.Sort(key=FreedomIndex)\;
\For{(idx,$\sim$) $\in$ PriorityOrder}{
Check for collision in SuccSteps with all robots in PriorityOrder[0:idx]\;
\If{Collision Detected}{
\tryy{\try}{
g[idx'] = g[idx]\; 
rhs[idx'] = rhs[idx]\; 
U[idx'] = U[idx]\;
c[idx'](u,v) = c[idx](u,v) $\forall (u,v) \in S$\;
c[idx']($s[idx]_{start}$,SuccSteps[idx]) = $\infty$\;
\updatevertex{idx',SuccSteps[idx]}\;
\compsp{idx'}\;
SuccSteps[idx] = arg$min_{s'\in Succ(s[idx]_{start})}(h(Z))$\;
}
\exc{\except {No Path}}{
SuccSteps[idx] = Previous Position of $R_{idx}$\;
}}}
\KwRet SuccSteps}

\SetKwFunction{compf}{ComputeFreedom}
\SetKwProg{CF}{Function}{:}{}
\CF{\compf{i,u}}{
freedom\_index = 0\;
\For{$s\in Succ(u)$}{\If{$c[i](u,s)\neq \infty$}{$freedom\_index$+=1 \;}}
\KwRet freedom\_index
}

\SetKwFunction{main}{Main}
\SetKwProg{MN}{Function}{:}{}
\MN{\main{}}{
\For{$i$ $\in$ robots}{
$s_{last}[i] = s_{start}[i]$\;
\init{i}\;
\compsp{i}\;
}
\While {$s_{start}[i] \neq s[i]_{goal}$ $\forall$ i in robots}{ 
SuccSteps = $\emptyset$\;
PriorityOrder = $\emptyset$\;
\For{$i$ $\in$ robots}{
\If{$s_{start}[i] \neq s[i]_{goal}$}{\If{$v_{obs}[i] \neq 0$ and $g[i](s_{start}[i]) \neq rhs[i](s_{start}[i])$}{UpdateVertex(i,$s_{start}[i]$)\;
ComputeShortestPath(i)\;}
if ($g[i](s_{start}[i]=\infty$): then there is no known path\;
SuccSteps.Append(arg $min_{s'\in Succ(s_{start}[i])}(c[i](s_{start}[i],s')+g[i](s'))$\;
PriorityOrder.Append([i,\compf{$s_{start}[i]$}])}
\Else{
SuccSteps.Append($s[i]_{goal}$)\;
PriorityOrder.Append(i,MaxFreedom)\;
}}
SuccSteps = ResolveConflicts(SuccSteps,PriorityOrder)\;
$s_{start}[i]$ = SuccSteps[i] $\forall \hspace{1ex} i \in$ robots\;
Move all robots to respective $s_{start}$\;
\For{$i$ $\in$ robots}{
Scan graph for changed edge costs\;
\If {any edge costs changed}{
$k_m[i] = k_m[i] + h(s_{last}[i],s_{start}[i])$\;
$s_{last}[i]=s_{start}[i]$\;
\For{all directed edges $(u,v)$ with changed edge costs}{
Update the edge cost $c[i](u,v)$\;
\updatevertex{i,u}\;}
\compsp{i}\;}
}}}
\nonl where, $Z = [s,s',idx,idx']$\\ 
\nonl and, $h(Z)= (c[idx'](s[idx]_{start},s')+g[idx'](s'))$
\end{multicols}
\vspace{3ex}
\caption{Proposed PD* algorithm}
\label{main_algo}
\end{algorithm*}

The planning of the next step happens in two stages. First, using D* Lite the next step for each robot is calculated without accounting for positions of the other robots in the environment. In the second stage, collisions between robots wanting to occupy the same node are handled in accordance with the priority list generated by the freedom index.

Algorithm \ref{conflict_res} shows this process in detail. The algorithm iterates over the priority list sequentially deconflicting the paths of each robot. $SuccSteps$ is a list containing the next steps (coordinates) for all the robots in sequential order i.e., $SuccSteps[i]$ contains the coordinates of the next step for robot $R_i$. For robot $R_{idx}$ the algorithm checks for collisions with all robots that have a higher priority than $R_{idx}$. If a collision is detected, the algorithm creates a copy of the graph and the D* Lite heuristics corresponding to $R_{idx}$. This is represented in lines 7--10. Then, the algorithm updates the cost of moving from $R_{idx}$’s current position to the proposed new position, to infinity. This effectively adds a virtual obstacle at this proposed new position forcing D* Lite to compute an alternative path to the target. D* Lite then updates the heuristics to accommodate the changed edge cost and the deconflicted next step for $R_{idx}$ is calculated and stored in $SuccSteps$.

To deal with contingencies that result from the added virtual obstacle completely blocking the path of $R_{idx}$, a backup mechanism has also been implemented (lines 15--16). In case the virtual obstacle blocks the path of the robot, then the new copy of the graph that has been created with the virtual obstacle is deleted and, in its place, the robot is instructed to move to its previously occupied position.

\subsection{PD* algorithm}
 Algorithm 4 shows the complete algorithm addressing the MRS path planning problem. The D* Lite algorithm (Algorithm \ref{ogD*}) was modified to incorporate multi-robot capabilities and the prioritization scheme (Algorithm \ref{freedom_index_algo}). Preserving the central D* Lite structure, its functions have been updated with the robot index input to ensure that the correct parameters are being calculated for the correct robot. Further, the main function was adapted in accordance with these changes to include multi-robot support with one step calculation and collision avoidance (Algorithm \ref{conflict_res}). The final algorithm now initializes the fields and calculates the initial heuristics for all robots, $R_i$. Then it moves each robot one step, and ensures a collision-free operation at each step. The process is repeated until all the robots have reached the target.

\section{Simulations and Discussions}
The PD* algorithm has been tested thoroughly through extensive simulation runs. The primary questions that are addressed through these results are: (1) the quality of the solution, and (2) the computational efficiency of the proposed method. The quality of the solution is measured in terms of the combined path length taken by the robots to reach the target. The computational efficiency is measured in terms of the time taken by the algorithm to reach completion. 

All experiments were run in python on a laptop with an i7-7700HQ CPU. The base implementation of D* Lite in python \cite{SamD*} was redesigned by integrating the freedom index based priority (Algorithms \ref{conflict_res} and \ref{freedom_index_algo}) scheme to handle MRS path planning.

\subsection{Simulation Environment}
For simulations, a 100$\times$100 unit 4-connected grid world was constructed with the Goal located at the center of the grid. Obstacle locations and robot start configurations were randomly generated with the following constraints:
\begin{itemize}
    \item The starting position of any robot cannot coincide with any obstacle.
    \item There exists at least one path for each robot from its starting configuration to the Goal in the generated configuration.
\end{itemize}

\begin{figure}[h!]
    \centering
    \includegraphics[width=\linewidth]{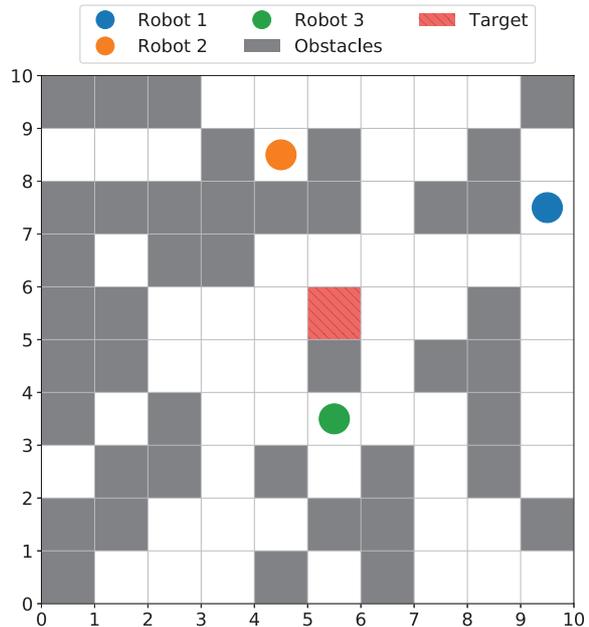}
    \caption{Sample 10$\times$10 environment with 3 robots}
    \label{fig:example_grid}
\end{figure}

Figure \ref{fig:example_grid} shows a sample 10$\times$10 grid world with three robots created using this methodology.

\begin{figure*}[h!]
    \centering
    \includegraphics[width=\linewidth]{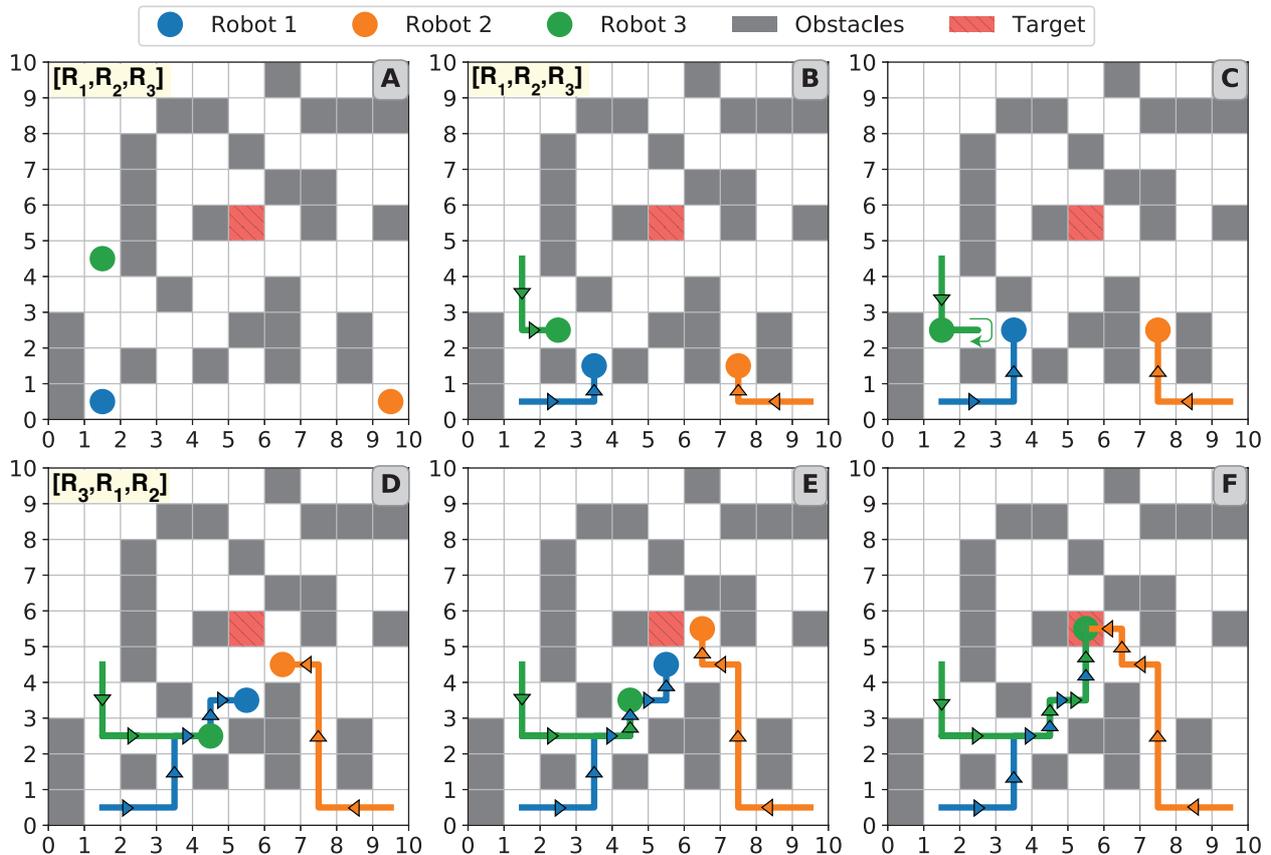}
    \caption{Conflict resolution using freedom index approach: (A). Initial configuration, (B). Configuration showing impending collision between $R_1$ and $R_3$, (C). Configuration showing mitigated collision, (D). Configuration showing impending collision between $R_1$ and $R_2$, (E). Configuration showing mitigated collision, (F). Final configuration after reaching target.}
    \label{fig:explanation}
\end{figure*}

Figure \ref{fig:explanation} demonstrates an example 10$\times$10 grid world with three robots showcasing the freedom index in action. Figure \ref{fig:explanation} (A) shows the initial configuration of the three robots. Robot 1 ($R_1$) is located at coordinate $(1, 0)$, Robot 2 ($R_2$) is located at $(9, 0)$ and Robot 3 ($R_3$) is located at $(1, 4)$. $R_1$, $R_2$ and $R_3$ have two, two and three possible path choices, respectively. Thus the priority list at this point is: [$R_1$, $R_2$, $R_3$] (also shown in the top left corner of the figure).

The algorithm then starts moving the robots along the calculated paths. The first possible collision is detected at the condition shown in Figure \ref{fig:explanation} (B). $R_1$ is at $(3, 1)$, $R_2$ is at $(7, 1)$ and $R_3$ is at $(2, 2)$. The predicted collision is between $R_1$ and $R_3$ as both robots wish to occupy $(3, 2)$ in the next time step. In this configuration, $R_1$, $R_2$ and $R_3$ have two, two and three paths respectively and thus the priority list at this point is: [$R_1$, $R_2$, $R_3$]. Hence, $R_1$ gets priority over $R_3$ to plan its paths. $R_1$ will occupy $(3, 2)$ in the next time step while $R_3$ will have to take a detour to accommodate $R_1$. The resultant configuration obtained in the next time step is seen in Figure \ref{fig:explanation} (C) where $R_1$ occupies $(3, 2)$ and $R_3$ occupies $(1,2)$.

A second collision is predicted in the configuration shown by Figure \ref{fig:explanation}(D).  $R_1$ is at $(5, 3)$, $R_2$ is at $(6, 4)$ and $R_3$ is at $(4, 2)$. In this case, the predicted collision is between $R_1$ and $R_2$ vying for $(5, 4)$ in the next time step. In this configuration, $R_1$, $R_2$ and $R_3$ have two, three and two paths respectively and thus the priority list at this point changes to: [$R_3$, $R_1$, $R_2$]. This allows $R_1$ to prioritize its path over $R_2$. In the next time step $R_1$ will occupy $(5, 4)$ while $R_2$ will have to take a detour to accommodate $R_1$. The resultant configuration obtained in the next time step is seen in Figure \ref{fig:explanation} (E) where $R_1$ occupies $(5, 4)$ and $R_2$ occupies $(6, 5)$. Figure \ref{fig:explanation} (F) shows the final configuration of this environment when all the robots have reached the targets along with the paths taken by the robots to reach the target. The priority orders for cases (C) and (E) are not mentioned because these states are after the collision and thus the priority order does not affect the next step of the robots. Similarly the priority order for case (F) is not mentioned as it shows the final configuration of the system.

\begin{figure*}[!b]
    \centering
    \includegraphics[width=\linewidth]{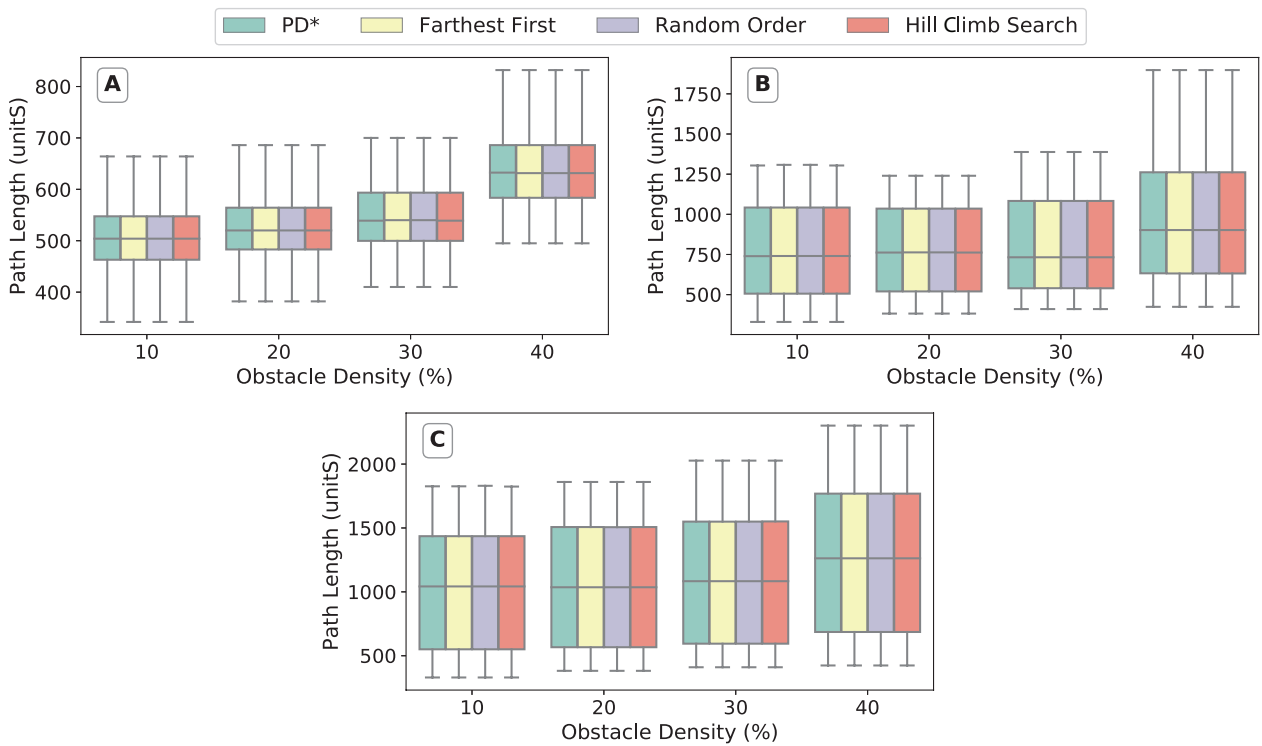}
    \caption{Comparison of mean path lengths over 100 simulation runs with (A) 10 robots, (B) 20 robots , and (C) 30 robots}
    \label{fig:length}
\end{figure*}

\subsection{Testing Methodology}
To understand the effects of the number of obstacles and number of robots the proposed PD* algorithm was tested in worlds that were 10\%, 20\%, 30\%, and 40\% filled with obstacles, with three sub cases each of 10, 20, and 30 robots, respectively. 100 worlds were randomly generated following the constraints mentioned above for a total of 1200 cases.

The performance of the proposed PD* algorithm was compared with 3 other methods: (I) Hill Climb Searching \cite{bennewitz2002finding}, (II) Farthest First \cite{van2005prioritized}, (III) Random Order (where the priority to each robot was arbitrarily assigned).

\subsection{Results}
\begin{figure*}[h!]
    \centering
    \includegraphics[width=\linewidth]{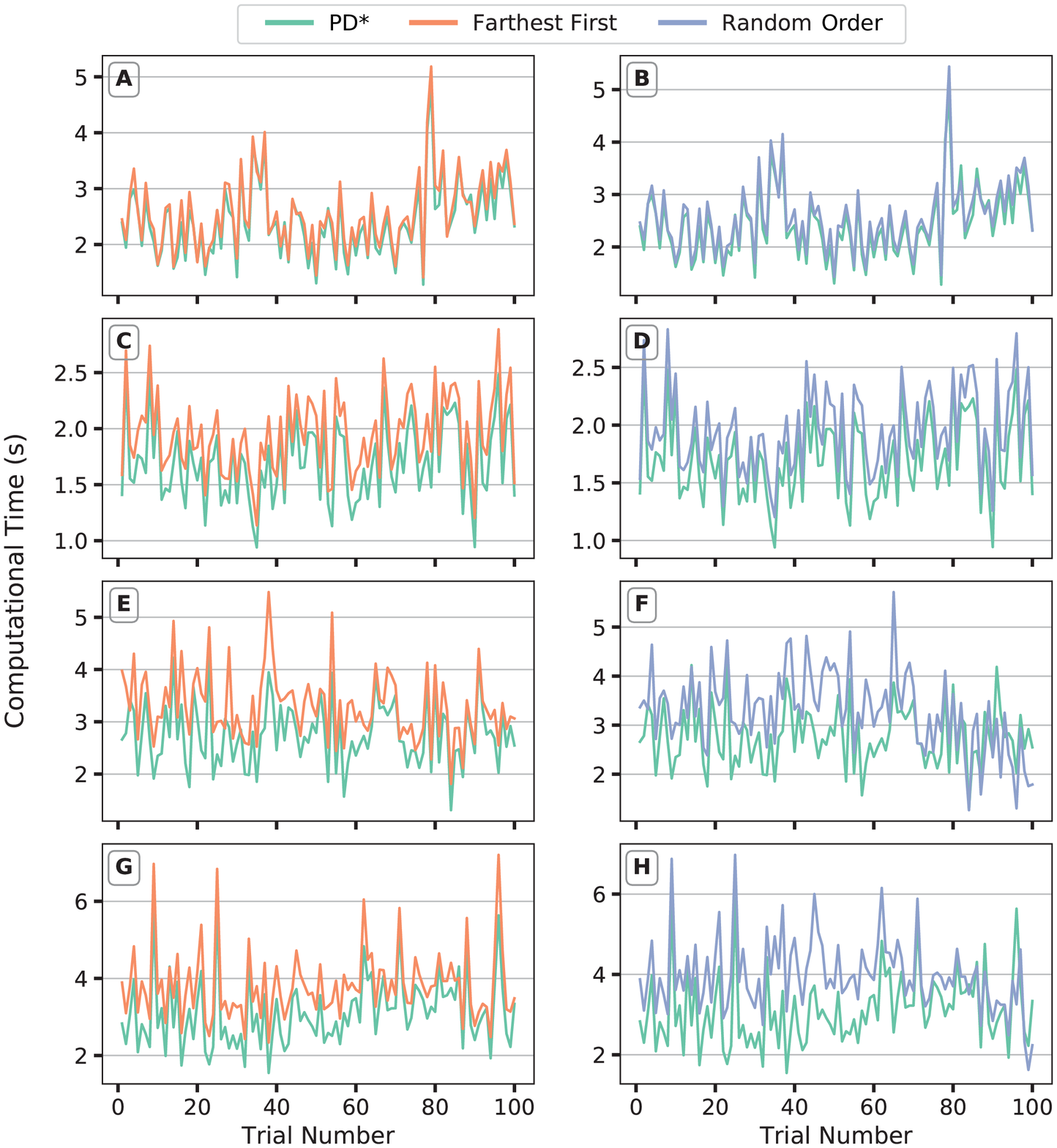}
    \caption{Comparison of computational time over 100 simulation runs with 20 robots; Plots (A), (C), (E), (F) - PD* against Farthest First with 10\%, 20\%, 30\% and 40\% obstacle density, respectively; Plots (B), (D), (F), (G) - PD* against Random Order with 10\%, 20\%, 30\% and 40\% obstacle density, respectively}
    \label{fig:individual_runs}
\end{figure*}

\begin{figure*}[!h]
    \centering
    \includegraphics[width=\linewidth]{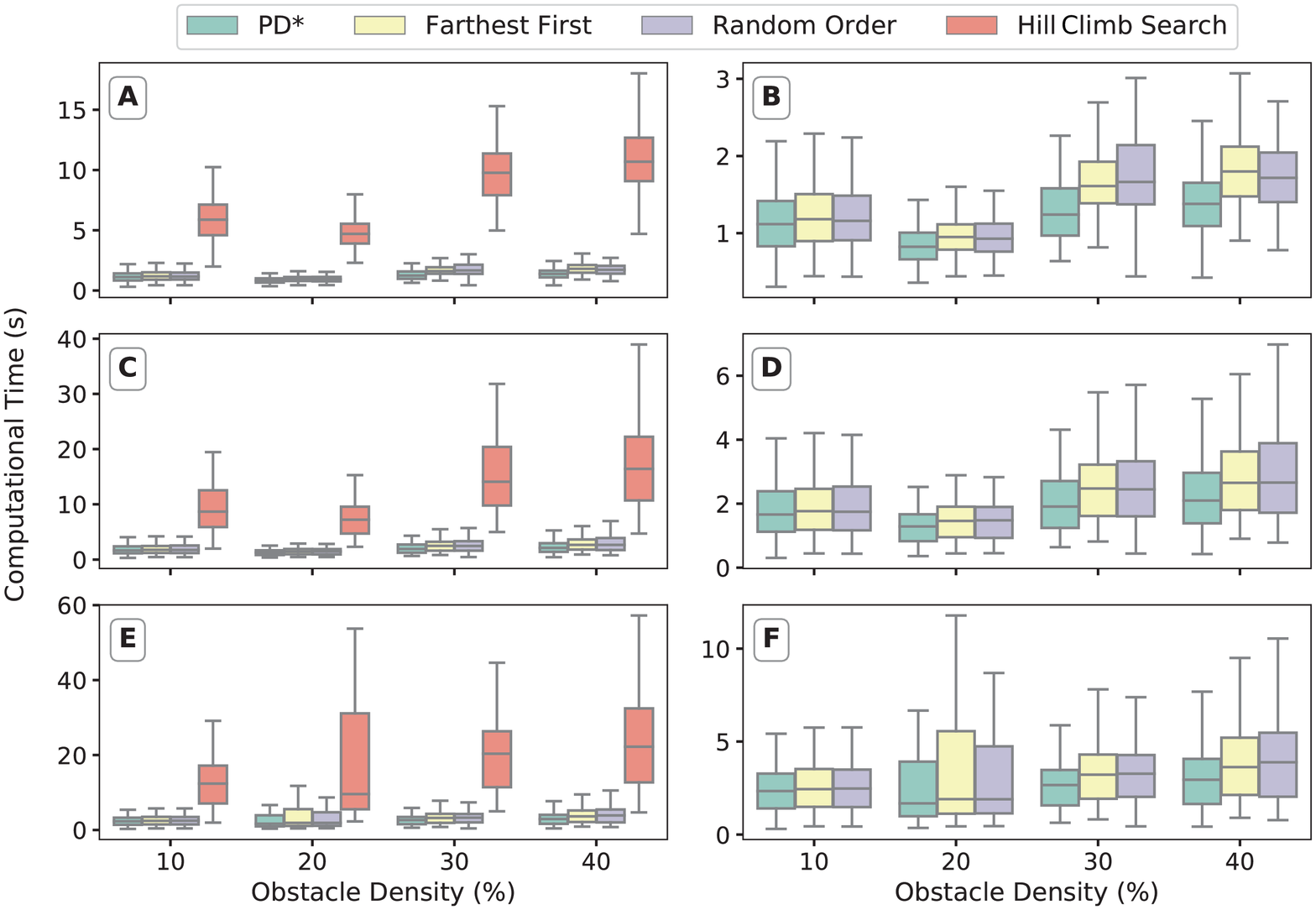}
    \caption{Comparison of mean computational time: Plots (A), (C), and (E) showing 10, 20, and 30 robots scenario, respectively; Plots (B), (D), and (F) showing 10, 20, and 30 robots scenario, excluding the Hill Climb Search method, respectively}
    \label{fig:time}
\end{figure*}

\begin{figure}[h!]
    \centering
    \includegraphics[width=\linewidth]{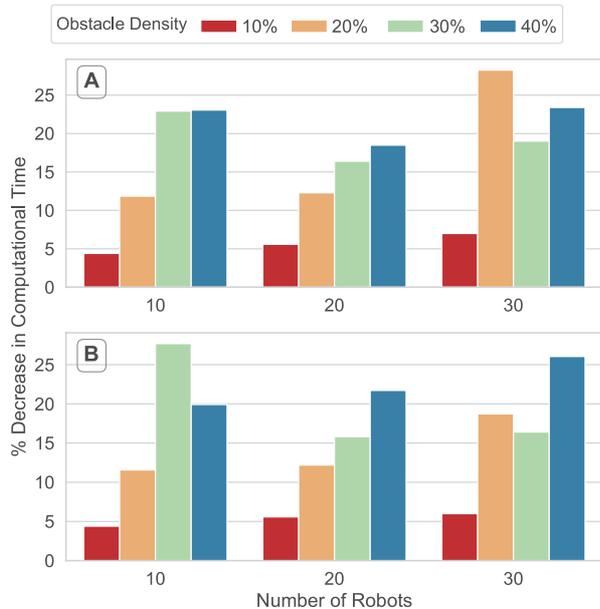}
    \caption{Percentage improvement in mean computational time: (A): Compared to Farthest First priority (B): Compared to Random Order priority.}
    \label{fig:efficiency}
\end{figure}

 The comparison results are shown in Figures \ref{fig:length}, \ref{fig:time}, and \ref{fig:efficiency}. Figure \ref{fig:length} shows the mean combined path length of all robots over the 100 runs for 10, 20, and 30 robot scenarios, respectively. The results show that the path lengths as computed by the proposed PD* algorithm are much similar, if not the same, as computed by the compared approaches. 

This is attributed to the use of D* Lite algorithm as the base path planning algorithm, and thus employing the same path-finding strategy. Furthermore, since the robots  behave as dynamic virtual obstacles, they are only present for one time step to avoid a collision. Hence, once a collision is mitigated D* Lite is allowed to reuse the original path. The results indicate that D* Lite indeed prefers the older path instead of taking a detour. Moreover, similar path lengths implies that the proposed PD* algorithm generates similar quality solutions compared to the existing ones.

Figure \ref{fig:individual_runs} compares the performance of the proposed PD* algorithm to the Farthest First and Random Order algorithms in terms of computational performance. The simulations were run on a twenty robot scenario for different obstacle densities. The Hill Climb Search method was not included in these comparisons as it takes an order of magnitude larger time for computation. This is because the algorithm solves each graph multiple times to find the best priority order. The left column in Figure \ref{fig:individual_runs} (plots (A),(C),(E),(F)) shows the performance of the proposed PD* algorithm compared to Farthest First algorithm in cases of 10\%, 20\%, 30\% and 40\% obstacle density, respectively. Similarly, the right column (plots (B), (D), (F), (G)) show the performance of the proposed PD* algorithm to Random Order algorithm. 

From Figure \ref{fig:individual_runs}, it is clear that the proposed PD* algorithm is computationally faster than the compared approaches in most cases. In the case of 10\% obstacle density the algorithm performs better in 89\% of the cases when compared to the Farthest First algorithm, and 90\% of the cases when compared to Random Order algorithm. In the case of 20\% obstacle density the PD* algorithm out performs both of the compared approaches in all cases. When the environment has 30\% obstacle density, the use of PD* results in reduced computational time in 97\% and 80\% cases when compared to Farthest First and Random Order algorithms, respectively, and in the case of 40\% density the efficiency improvements are found to be 90\% and 93\%, respectively. Furthermore, it can be inferred from the plots that the difference in computational time increases with the increase in obstacle density. For brevity, plots for simulations scenarios with 10 and 30 robots have been omitted. However, to analyse the performance of the algorithm in all the aforementioned scenarios, comparison plots presenting the mean computation time are presented next.


Figure \ref{fig:time} shows the mean computational time required over the 100 runs. Unlike the path lengths, the results indicate that the PD* algorithm is significantly more efficient than the competing algorithms in almost all cases. The Hill Climb Search method takes substantially longer time as seen in Figures \ref{fig:time} (A), (C), and (E). Figures \ref{fig:time} (B), (D), and (F) presents the comparison of the PD* algorithm with the Farthest First and Random Order algorithms, where plots presented in Figures \ref{fig:time} (A), (C), and (E) have been reproduced excluding the Hill Climb Search algorithm for better clarity. 

From these plots, it can be clearly seen that the difference in computational time between the proposed PD* algorithm and the compared approaches is larger in densely populated environments or environments with many robotic agents. In the case of ten robots (Figure \ref{fig:time}(B)), the proposed PD* algorithm takes on average around 0.3 seconds less to solve the problem than the next best approach. The first quartile of the results above the mean are also lower than the mean values of the compared algorithm in most cases. In the case of twenty bots (Figure \ref{fig:time}(D)) the PD* algorithm finishes 0.5 seconds faster with upper quartile lower than the upper quartile of compared algorithms. Similarly, in the case of thirty bots (Figure \ref{fig:time}(F)) the proposed PD* algorithm takes approximately 1 second less than the other approaches with the upper quartile being lower than corresponding upper quartiles.

Figure \ref{fig:efficiency} quantifies the efficiency increase (in terms of decrease in computational time required) provided by the proposed PD* algorithm compared to algorithms (II) and (III). The Hill Climb Search method was excluded from this comparison as well due to same reasons. The proposed PD* algorithm was found to be approximately 5\% more efficient compared to other algorithms in sparsely populated environments. Interestingly, the computational efficiency was found to be increasing with the increase in obstacle density. In environments that are 20\% filled, the proposed PD* algorithm provides over 10\% efficiency increase in all cases with an efficiency increase of almost 25\% in the case of 30 robots. Similarly, the proposed PD* algorithm achieves efficiency increase of over 15\% in environments that are 30\% filled, and over 20\% in environments that are 40\% filled.

This trend is significant because in dense environments or in environments with a large number of robots, the probability of collisions are much higher, demanding a more efficacious approach. From the discussions presented above, the proposed PD* algorithm is found to be more effective in such environments demonstrating its applicability to problems involving densely populated environments or environments with many robotic agents.

\section{Conclusions}
In this paper a generally applicable prioritized path planning method, PD* is introduced for MRS. Using extensive computer simulations it is shown that the proposed method is significantly more efficient than state-of-the-art algorithms particularly with large number of robots in environments with higher obstacle densities while providing solutions of equivalent quality. Furthermore, it is shown that the PD* scales well with the number of robots and is sufficiently fast for practical applications.

In this paper the efficacy of PD* is demonstrated in 2D environments. However, it is trivial to extend it to 3D applications by simply changing the graph definition. Another important observation is the modularity provided by the method. While this paper utilizes the D* Lite as the base path planning algorithm, it can be easily replaced with any other path planning algorithm depending on the requirements of the task at hand.

Lastly, this paper assumed that all robots are equivalent in their geometry and performance, have a common goal/target, and start moving at the same time. However, with some minor changes, the proposed PD* algorithm would be valid without these assumptions and thus, can be applied in more generalized scenarios with heterogeneous agents.

A significant extension of this work involves implementing a queuing system in conjunction with the proposed method. In MRS with large number of robots, sizable queues might be created in environments with narrow passageways. It is interesting to investigate the effect on efficiency that a queuing system might have in such scenarios. Notions of Pareto optimality \cite{ghrist2004pareto} could be used in this regard. 

\section*{Acknowledgements}
The authors would like to thank Prof. Neeldhara Misra, of Indian Institute of Technology Gandhinagar, for the fruitful discussions and Indian Institute of Technology Gandhinagar for their support.

\bibliography{main}

\begin{thebibliography}{10}
\expandafter\ifx\csname url\endcsname\relax
  \def\url#1{\texttt{#1}}\fi
\expandafter\ifx\csname urlprefix\endcsname\relax\def\urlprefix{URL }\fi
\expandafter\ifx\csname href\endcsname\relax
  \def\href#1#2{#2} \def\path#1{#1}\fi

\bibitem{acemoglu2020robots}
D.~Acemoglu, P.~Restrepo, Robots and jobs: Evidence from us labor markets,
  Journal of Political Economy 128~(6) (2020) 2188--2244.

\bibitem{inproceedings}
P.~Grippa, D.~A. Behrens, C.~Bettstetter, F.~Wall, Job selection in a network
  of autonomous uavs for delivery of goods, arXiv preprint arXiv:1604.04180
  (2016).

\bibitem{enright_optimization}
J.~J. Enright, P.~R. Wurman, Optimization and coordinated autonomy in mobile
  fulfillment systems, in: Proceedings of the 9th AAAI Conference on Automated
  Action Planning for Autonomous Mobile Robots, AAAI Press, San Francisco, USA,
  2011, p. 33–38.

\bibitem{arbanas2018decentralized}
B.~Arbanas, A.~Ivanovic, M.~Car, M.~Orsag, T.~Petrovic, S.~Bogdan,
  Decentralized planning and control for uav--ugv cooperative teams, Autonomous
  Robots 42~(8) (2018) 1601--1618.

\bibitem{hyldmar2019fleet}
N.~Hyldmar, Y.~He, A.~Prorok, A fleet of miniature cars for experiments in
  cooperative driving, in: 2019 International Conference on Robotics and
  Automation (ICRA), IEEE, Montreal, Canada, 2019, pp. 3238--3244.

\bibitem{tsugawa_transport}
S.~{Tsugawa}, S.~{Kato}, T.~{Matsui}, H.~{Naganawa}, H.~{Fujii}, An
  architecture for cooperative driving of automated vehicles, in: 2000 IEEE
  Intelligent Transportation Systems Proceedings, Dearborn, USA, 2000, pp.
  422--427.

\bibitem{Pavone_load_balancing}
M.~Pavone, S.~L. Smith, E.~Frazzoli, D.~Rus, Robotic load balancing for
  mobility-on-demand systems, The International Journal of Robotics Research
  31~(7) (2012) 839--854.

\bibitem{parker2007distributed}
L.~E. Parker, Distributed intelligence: Overview of the field and its
  application in multi-robot systems., in: AAAI Fall Symposium: Regarding the
  Intelligence in Distributed Intelligent Systems, Arlington, USA, 2007, pp.
  1--6.

\bibitem{olcay2020collective}
E.~Olcay, F.~Schuhmann, B.~Lohmann, Collective navigation of a multi-robot
  system in an unknown environment, Robotics and Autonomous Systems 132 (2020)
  103604.

\bibitem{de2019bio}
J.~P. L.~S. de~Almeida, R.~T. Nakashima, F.~Neves-Jr, L.~V.~R. de~Arruda,
  Bio-inspired on-line path planner for cooperative exploration of unknown
  environment by a multi-robot system, Robotics and Autonomous Systems 112
  (2019) 32--48.

\bibitem{Geihs_2020}
K.~Geihs, Engineering challenges ahead for robot teamwork in dynamic
  environments, Applied Sciences 10~(4) (2020) 1368.

\bibitem{nilsson2014principles}
N.~J. Nilsson, Optimization and Coordinated Autonomy in Mobile Fulfillment
  Systems, Morgan Kaufmann, 2014.

\bibitem{koenig2002d}
S.~Koenig, M.~Likhachev, D\^{}* lite, Aaai/iaai 15 (2002) 476--483.

\bibitem{dijkstra1959note}
E.~W. Dijkstra, et~al., A note on two problems in connexion with graphs,
  Numerische mathematik 1~(1) (1959) 269--271.

\bibitem{moore1959shortest}
E.~F. Moore, The shortest path through a maze, in: Proc. Int. Symp. Switching
  Theory, 1959, Boston, USA, 1959, pp. 285--292.

\bibitem{tarjan1972depth}
R.~Tarjan, Depth-first search and linear graph algorithms, SIAM journal on
  computing 1~(2) (1972) 146--160.

\bibitem{ferguson2005guide}
D.~Ferguson, M.~Likhachev, A.~Stentz, A guide to heuristic-based path planning,
  in: Proceedings of the international workshop on planning under uncertainty
  for autonomous systems, international conference on automated planning and
  scheduling (ICAPS), Monterey, USA, 2005, pp. 9--18.

\bibitem{lavalle2006planning}
S.~M. LaValle, Planning algorithms, Cambridge university press, 2006.

\bibitem{barraquand1991robot}
J.~Barraquand, J.-C. Latombe, Robot motion planning: A distributed
  representation approach, The International Journal of Robotics Research
  10~(6) (1991) 628--649.

\bibitem{olmi2011efficient}
R.~Olmi, C.~Secchi, C.~Fantuzzi, An efficient control strategy for the traffic
  coordination of agvs, in: 2011 IEEE/RSJ International Conference on
  Intelligent Robots and Systems, IEEE, San Francisco, USA, 2011, pp.
  4615--4620.

\bibitem{draganjac2016decentralized}
I.~Draganjac, D.~Mikli{\'c}, Z.~Kova{\v{c}}i{\'c}, G.~Vasiljevi{\'c},
  S.~Bogdan, Decentralized control of multi-agv systems in autonomous
  warehousing applications, IEEE Transactions on Automation Science and
  Engineering 13~(4) (2016) 1433--1447.

\bibitem{ferner2013odrm}
C.~Ferner, G.~Wagner, H.~Choset, Odrm* optimal multirobot path planning in low
  dimensional search spaces, in: 2013 IEEE International Conference on Robotics
  and Automation, IEEE, Karlsruhe, Germany, 2013, pp. 3854--3859.

\bibitem{wagner2011m}
G.~Wagner, H.~Choset, M*: A complete multirobot path planning algorithm with
  performance bounds, in: 2011 IEEE/RSJ international conference on intelligent
  robots and systems, IEEE, San Francisco, USA, 2011, pp. 3260--3267.

\bibitem{van2005prioritized}
J.~P. Van Den~Berg, M.~H. Overmars, Prioritized motion planning for multiple
  robots, in: 2005 IEEE/RSJ International Conference on Intelligent Robots and
  Systems, IEEE, Edmonton, Canada, 2005, pp. 430--435.

\bibitem{jose2016task}
K.~Jose, D.~K. Pratihar, Task allocation and collision-free path planning of
  centralized multi-robots system for industrial plant inspection using
  heuristic methods, Robotics and Autonomous Systems 80 (2016) 34--42.

\bibitem{bennewitz2002finding}
M.~Bennewitz, W.~Burgard, S.~Thrun, Finding and optimizing solvable priority
  schemes for decoupled path planning techniques for teams of mobile robots,
  Robotics and autonomous systems 41~(2-3) (2002) 89--99.

\bibitem{zimmer2020adaptive}
C.~Zimmer, D.~Driess, M.~Meister, N.-T. Duy, Adaptive discretization for
  evaluation of probabilistic cost functions, in: International Conference on
  Artificial Intelligence and Statistics, Virtual Conference, 2020, pp.
  2098--2108.

\bibitem{lozano}
T.~Lozano-Perez, Spatial planning: A configuration space approach, IEEE
  Transactions on Computers 32~(02) (1983) 108--120.

\bibitem{henrichcspace}
D.~{Henrich}, C.~{Wurll}, H.~{Worn}, Online path planning with optimal c-space
  discretization, in: Proceedings. 1998 IEEE/RSJ International Conference on
  Intelligent Robots and Systems, Vol.~3, Victoria, Canada, 1998, pp.
  1479--1484 vol.3.

\bibitem{erssonunknown}
T.~{Ersson}, {Xiaoming Hu}, Path planning and navigation of mobile robots in
  unknown environments, in: Proceedings 2001 IEEE/RSJ International Conference
  on Intelligent Robots and Systems, Vol.~2, Maui USA, 2001, pp. 858--864
  vol.2.

\bibitem{stentz1997optimal}
A.~Stentz, Optimal and efficient path planning for partially known
  environments, in: Intelligent unmanned ground vehicles, Springer, 1997, pp.
  203--220.

\bibitem{D*}
A.~Stentz, The focussed d* algorithm for real-time replanning, in: Proceedings
  of the 14th International Joint Conference on Artificial Intelligence -
  Volume 2, Montreal, Canada, 1995, p. 1652–1659.

\bibitem{ARA*}
M.~Likhachev, D.~I. Ferguson, G.~J. Gordon, A.~Stentz, S.~Thrun, Anytime
  dynamic a*: An anytime, replanning algorithm., in: International Conference
  on Automated Planning \& Scheduling, Vol.~5, Monterey, USA, 2005, pp.
  262--271.

\bibitem{setiawan2014experimental}
Y.~D. Setiawan, P.~S. Pratama, S.~K. Jeong, V.~H. Duy, S.~B. Kim, Experimental
  comparison of a* and d* lite path planning algorithms for differential drive
  automated guided vehicle, in: AETA 2013: Recent Advances in Electrical
  Engineering and Related Sciences, Springer, 2014, pp. 555--564.

\bibitem{Rosa_Comp}
W.~C. da~Rosa, I.~V. de~Bessa, L.~C. Cordeiro, Application of global
  route-planning algorithms with geodesy, arXiv preprint arXiv:1610.04597
  (2016).

\bibitem{SamD*}
S.~Sam, pydstarlite, \url{\\https://github.com/samdjstephens/pydstarlite} (2018
  (accessed Aug. 2020)).

\bibitem{ghrist2004pareto}
R.~Ghrist, J.~M. O’Kane, S.~M. LaValle, Pareto optimal coordination on
  roadmaps, in: Algorithmic foundations of robotics VI, Springer, 2004, pp.
  171--186.

\end{thebibliography}
\balance
\end{document}